%% file: main.tex
\documentclass[10pt,twocolumn,letterpaper]{article}

\usepackage{cvpr}      % To produce the REVIEW version

\usepackage{indentfirst} 

\definecolor{cvprblue}{rgb}{0.21,0.49,0.74}
\usepackage[pagebackref,breaklinks,colorlinks,allcolors=cvprblue]{hyperref}

\def\paperID{*****} % *** Enter the Paper ID here
\def\confName{CVPR}
\def\confYear{2025}

\title{MASSeg : 2nd Technical Report for 4th PVUW MOSE Track} % test-title

\author{
Xuqiang Cao\textsuperscript{1} \quad
Linnan Zhao\textsuperscript{1} \quad
Jiaxuan Zhao\textsuperscript{1} \quad
Fang Liu\textsuperscript{1} \\
Puhua Chen\textsuperscript{1*} \quad
Wenping Ma\textsuperscript{1} \\
\\
\textsuperscript{1}Key Laboratory of Intelligent Perception and Image Understanding \\ Xi'an, China
}

\begin{document}
\maketitle
\input{sec/0_abstract}    
\input{sec/1_intro}

\input{sec/2_formatting}
\input{sec/3_finalcopy}
{
    \small
    \bibliographystyle{ieee_fullname}
    \bibliography{main}
}

% WARNING: do not forget to delete the supplementary pages from your submission 
% \input{sec/X_suppl}

\end{document}

%% file: sec/0_abstract.tex
\begin{abstract}
\indent Complex video object segmentation continues to face significant challenges in small object recognition, occlusion handling, and dynamic scene modeling. This report presents our solution, which ranked \textbf{second} in the \textit{MOSE track of CVPR 2025 PVUW Challenge}. Based on an existing segmentation framework, we propose an improved model named \textbf{MASSeg} for complex video object segmentation, and construct an enhanced dataset, \textbf{MOSE+}, which includes typical scenarios with occlusions, cluttered backgrounds, and small target instances. During training, we incorporate a combination of inter-frame \textit{consistent} and \textit{inconsistent} data augmentation strategies to improve robustness and generalization. During inference, we design a \textit{mask output scaling strategy} to better adapt to varying object sizes and occlusion levels. As a result, MASSeg achieves a \textbf{J score of 0.8250}, \textbf{F score of 0.9007}, and a \textbf{J\&F score of 0.8628} on the MOSE test set.The code is available at \url{https://github.com/cxqNet/MASSeg}.
\end{abstract}

%% file: sec/1_intro.tex
\section{Introduction}
\label{sec:intro}

\indent Pixel-level scene understanding is a fundamental task in computer vision, aiming to recognize the category, semantics, and instance of each pixel. With the rapid growth of video content, this task has extended from static images to dynamic videos, drawing increasing attention to video object segmentation (VOS). VOS requires segmenting and tracking target objects across frames with only the first-frame mask provided. It has been widely applied in autonomous driving, augmented reality, video editing, and data annotation.

As task complexity increases, memory-based VOS methods have become the mainstream~\cite{ding2020semantic}. These models maintain a memory bank to store visual features and adopt attention or feature matching to segment targets in future frames. Representative works include STM~\cite{Oh2019STM}, which introduces

\begin{figure}[ht]
  \centering
  \includegraphics[width=\linewidth]{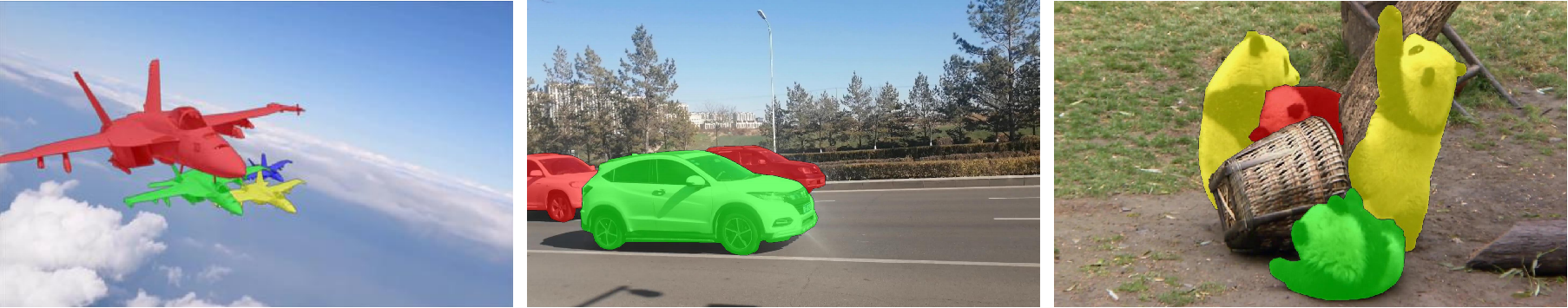}
  \caption{Representative challenges in complex video object segmentation. The examples showcase small object motion (left), appearance confusion with occlusion (middle), and densely cluttered scenes (right), which reflect typical situations encountered in the MOSE dataset.}
  \label{fig:challenges}
\end{figure}

\noindent explicit memory for pixel-level matching; STCN~\cite{Cheng2021STCN}, which improves efficiency with unmasked frames and L2 affinity; and XMem~\cite{Cheng2022XMem}, which mimics human memory with sensory, working, and long-term stages. Cutie~\cite{Tang2022Cutie} enhances robustness in crowded scenes via object-level memory and object transformers, achieving state-of-the-art results on YouTube-VOS and DAVIS.

To better evaluate model robustness in real-world scenarios, Henghui Ding et al. proposed and organized the MOSE~\cite{ding2024pvuw2024challengecomplex} and MeViS~\cite{MeViS} tracks. These tracks target two distinct but complementary aspects of video understanding: MOSE focuses on complex video object segmentation under challenging visual conditions, while MeViS emphasizes motion expression-guided segmentation based on natural language descriptions. 

To better evaluate model robustness in real-world scenarios, Henghui Ding et al., in collaboration with CVPR, introduced the Pixel-level Video Understanding in the Wild (PVUW) Challenge. The challenge focuses on pixel-level scene understanding and features two complementary tracks: MOSE~\cite{MOSE}, which targets complex multi-object segmentation under challenging visual conditions, and MeViS~\cite{MeViS}, which emphasizes motion expression-guided segmentation based on natural language descriptions. The MOSE track achieved significant progress in the PVUW 2024 Challenge~\cite{ding2024pvuw2024challengecomplex}.

\begin{figure*}[t]
  \centering
  \includegraphics[width=\textwidth]{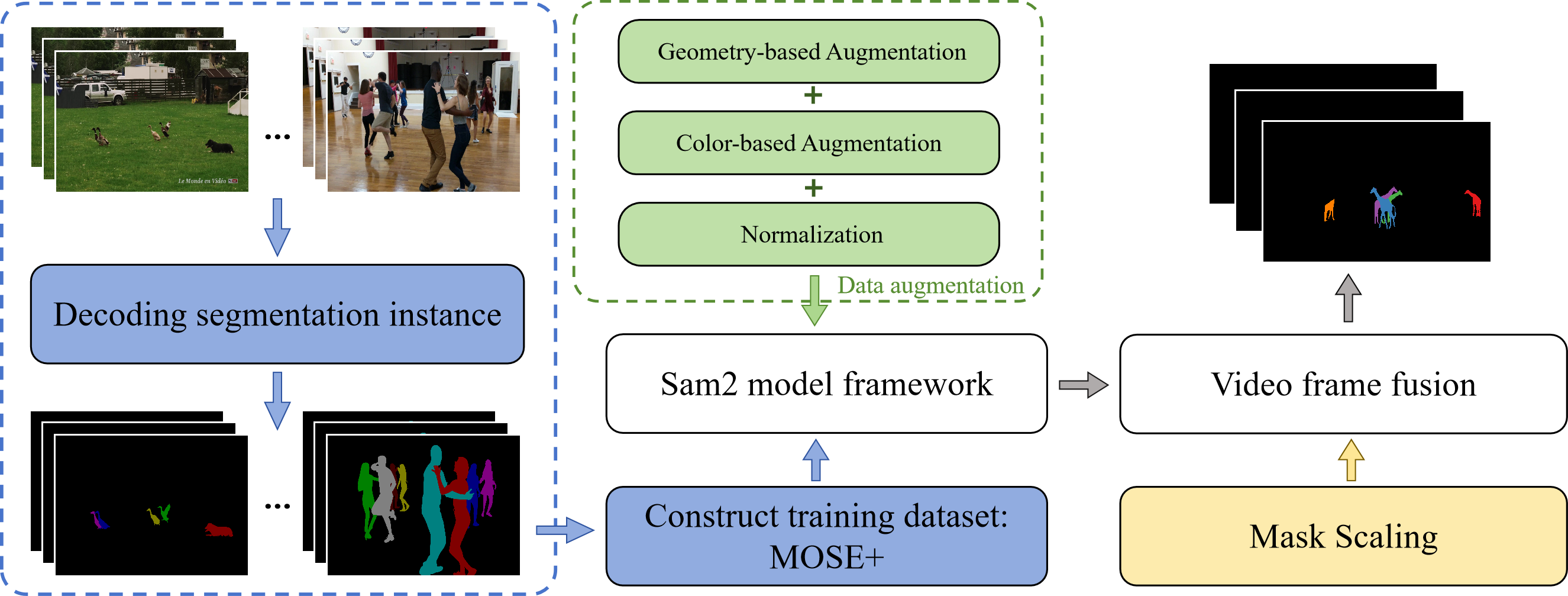}
  \caption{Overview of our method.}
  \label{fig:framework_all}
\end{figure*}

The MOSE dataset includes 2,149 videos and 5,200 annotated instances over 36 categories, with over 430K pixel-level masks. MOSE emphasizes real-world challenges such as occlusions, fast motion, dense small objects, similar appearances, and frequent reappearance, providing a rigorous benchmark beyond prior datasets like DAVIS and YouTube-VOS. As illustrated in Figure~\ref{fig:challenges}, the MOSE dataset contains challenging scenarios such as small objects, heavy occlusions, and similar instances, which significantly impact segmentation robustness.

Mainstream methods, e.g., Cutie, achieve over 84\% J\&F on YouTube-VOS but drop significantly on MOSE, revealing generalization limitations. MOSE has thus emerged as a valuable platform for driving progress in robust, real-world VOS.

To address MOSE-specific challenges, we propose an improved method based on the SAM2 framework~\cite{ravi2024sam2}, incorporating a mask output control strategy and retraining on an enhanced dataset \textbf{MOSE+}, which includes diverse small object instances and complex scenes. We further adopt inter-frame \textit{consistent} and \textit{inconsistent} augmentations and tuned inference strategies to improve robustness across scales and occlusions. Our method achieves a \textbf{J score of 0.8250}, \textbf{F score of 0.9007}, and a \textbf{J\&F of 0.8628} on the MOSE test set, ranking 2\textsuperscript{nd} overall.

%% file: sec/2_formatting.tex
\section{Method}
\label{sec:method}

Our overall solution is illustrated in Figure~\ref{fig:framework_all}. Based on the data characteristics of the MOSE dataset, we construct a customized dataset named \textbf{MOSE+} and design a series of targeted data augmentation strategies, aiming to simulate appearance variations, pose perturbations, illumination changes, and structural blur commonly found in real-world video scenarios. During inference, a \textit{mask confidence control strategy} is adopted, followed by a fusion process over video frames to obtain the final segmentation results. The detailed components are described below.

\subsection{Baseline Model}
\label{sec:baseline}

We adopt a transformer-based segmentation framework featuring object-guided attention, mask-aware memory, and spatiotemporal reasoning. The model effectively captures temporal cues and spatial details through dual memory modules and multi-scale decoding, enabling robust performance under challenging scenarios such as occlusion, motion blur, and small-object clutter. This strong baseline lays a solid foundation for our enhancement strategies.

\begin{figure*}[t]
  \centering
  \includegraphics[width=\textwidth]{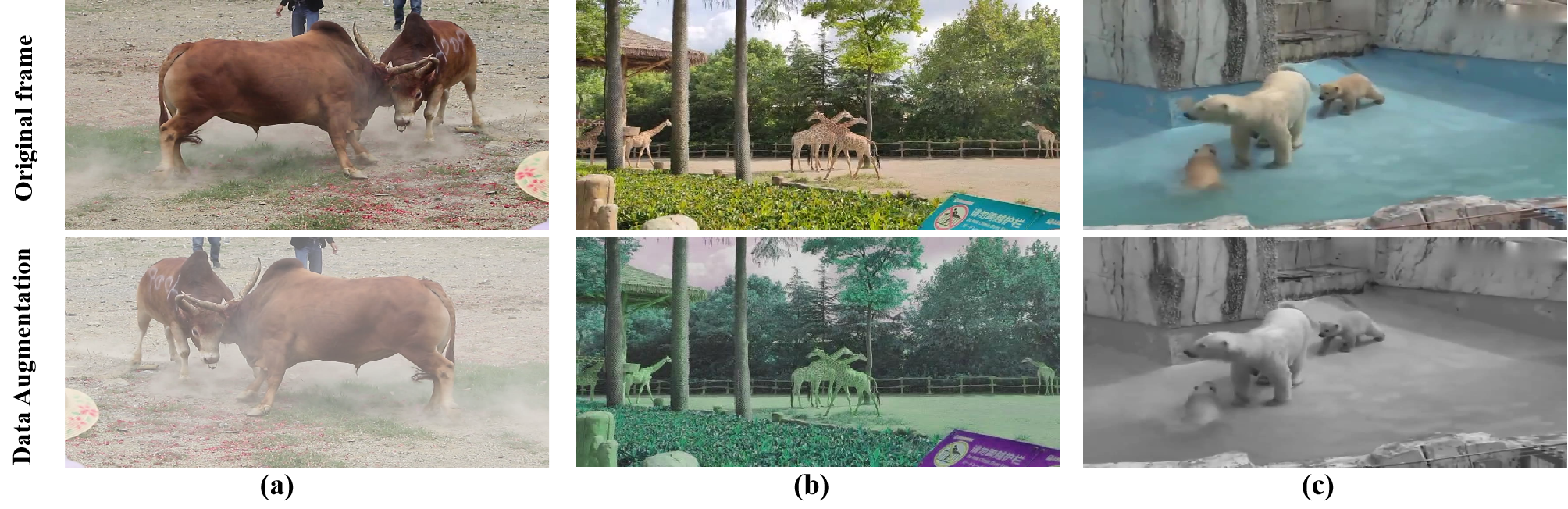}
  \caption{Visualization of our data augmentation strategies. Each column shows an example before (top) and after (bottom) applying augmentation. From left to right: (a) geometric transformation (e.g., affine distortion), (b) color jittering with inconsistent color shift, and (c) grayscale conversion. These augmentations simulate realistic variations in pose, illumination, and appearance, improving robustness and generalization of the model in complex scenarios.}
  \label{fig:data_enc}
\end{figure*}

\subsection{Loss Function}
\label{sec:loss}

To achieve high-precision segmentation and temporal consistency, we design a multi-task loss that combines pixel-wise accuracy, region-level overlap, classification discriminability, and robustness to occlusion. The total loss is defined as:

\begin{equation}
\mathcal{L}_{total} = \lambda_1 \mathcal{L}_{CE} + \lambda_2 \mathcal{L}_{Dice} + \lambda_3 \mathcal{L}_{Sim} + \lambda_4 \mathcal{L}_{MaskIoU}
\end{equation}

where $\mathcal{L}_{CE}$ denotes cross-entropy loss for foreground-background classification, $\mathcal{L}_{Dice}$ enhances region consistency, $\mathcal{L}_{Sim}$ enforces similarity between memory and query features, and $\mathcal{L}_{MaskIoU}$ constrains predicted mask quality. These losses are computed across multiple frames and candidate masks to jointly supervise spatiotemporal modeling.

In our implementation, the weights $\lambda_i$ are empirically set for balanced optimization. This formulation improves segmentation performance on small or occluded targets and supports temporal consistency in long sequences.

\subsection{Data Augmentation}
\label{sec:data_aug}

To improve generalization and robustness, we introduce a set of targeted augmentation strategies during training. Unlike static image tasks, video segmentation demands consistency across frames while simulating realistic variations. Our approach integrates both frame-consistent and frame-inconsistent perturbations:

\begin{itemize}
  \item \textbf{Consistent geometric transformations}: Random horizontal flipping, affine transformations (rotation, shear), and multi-scale resizing are applied across all frames in a clip to simulate viewpoint and object deformation.
  \item \textbf{Mixed color perturbations}: Brightness, contrast, and saturation changes are applied globally, while grayscale conversion and inconsistent color jittering are selectively applied to individual frames, enhancing robustness to lighting changes and visual ambiguity.
  \item \textbf{Normalization}: Images are transformed into tensors and normalized using ImageNet mean and standard deviation for stable convergence and pretrained compatibility.
\end{itemize}

These augmentations significantly improve the model's ability to handle structure variation, appearance change, and dynamic scenes in MOSE-like scenarios.

\subsection{Inference Strategy}
\label{sec:inference}

To enhance the robustness and adaptability of the model in complex video object segmentation, we design and apply a series of strategies during inference.

\textbf{Mask Confidence Control Strategy.} We observe that the quality of predicted masks can be significantly affected by post-processing in different scenarios, such as small objects, heavy occlusions, and target overlaps. To address this, we adopt a control strategy based on dynamic adjustment of the mask output distribution, using two key parameters: \textit{sigmoid scale} and \textit{sigmoid bias}. The sigmoid scale controls the sharpness of the output boundaries, while the sigmoid bias 

\begin{figure*}[t]
  \centering
  \includegraphics[width=\textwidth]{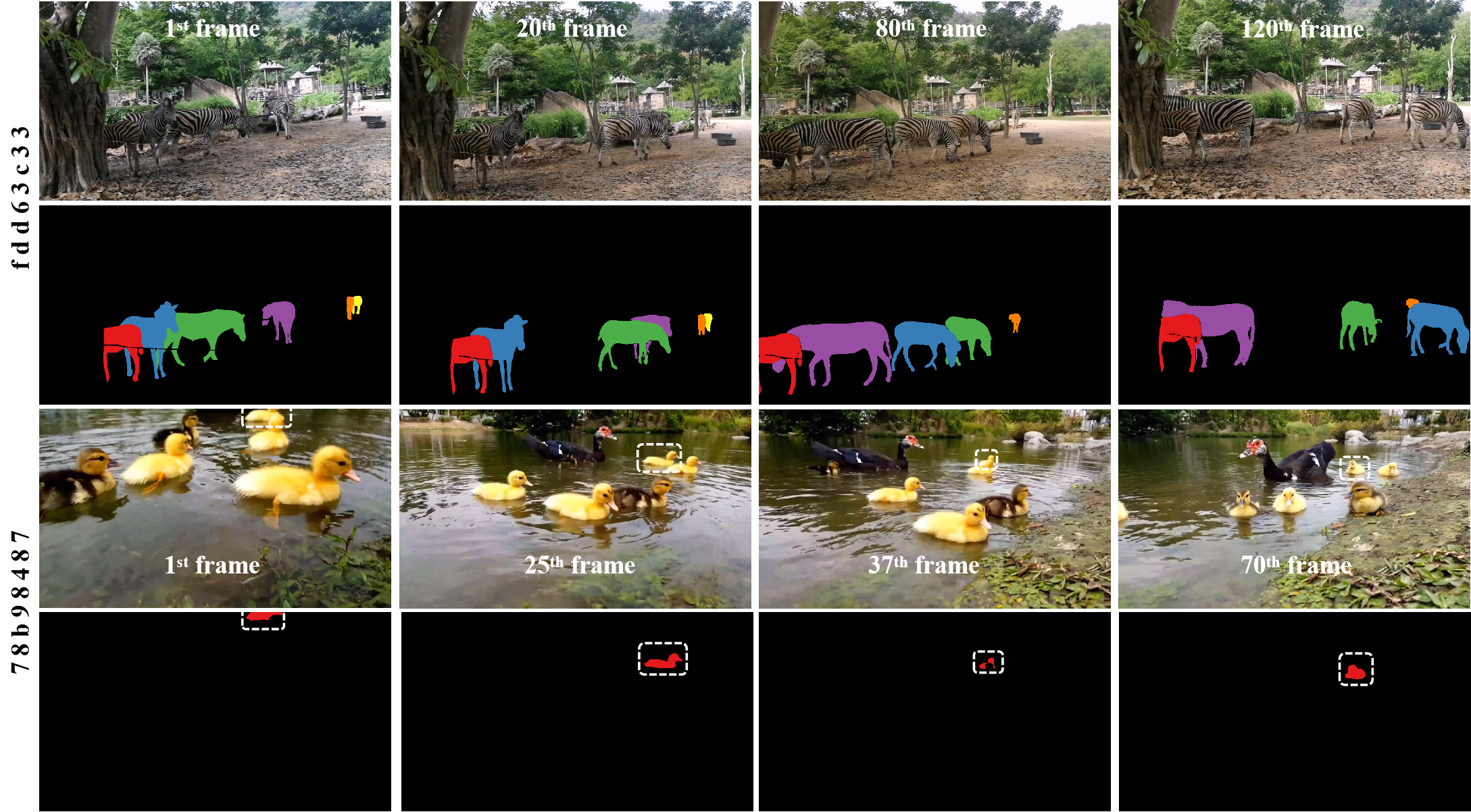}
  \caption{Qualitative results of our method on challenging MOSE test sequences. Our model accurately segments small objects, handles severe occlusions, and maintains temporal consistency across fast-moving and cluttered scenes.}
  \label{fig:output}
\end{figure*}

\noindent adjusts the overall activation level, thereby influencing the target coverage and boundary quality.

Extensive experiments on the validation set show that setting the sigmoid scale to 7.5 and the sigmoid bias to -4.0 yields the best performance on the MOSE dataset in terms of J\&F metrics. In addition, other parameter combinations are explored to improve local segmentation performance under specific conditions such as dense small objects or complex backgrounds. Final predictions are obtained by merging results from different local configurations to achieve globally optimal segmentation.

%% file: sec/3_finalcopy.tex
\section{Implementation Details}
\label{sec:implementation}

\textbf{Data.} To improve generalization and target modeling in complex scenarios, we construct an enhanced training set named \textbf{MOSE+}, based on the original MOSE dataset. This augmented set is composed of video segments from multiple public VOS datasets, selected to match the characteristics of MOSE, including frequent occlusions, dense small objects, object reappearance, and high similarity among targets. Specifically, we integrate carefully chosen sequences from datasets such as BURST~\cite{athar2023burst}, DAVIS~\cite{Pont-Tuset_arXiv_2017}, OVIS~\cite{qi2022occluded}, and YouTubeVIS~\cite{vos2019}, unify their annotations and resolution formats, and seamlessly merge them with MOSE to form a consistent training set, enhancing semantic understanding and robustness.

\textbf{Training.} We train our model end-to-end on the MOSE+ dataset. Each batch contains 2 samples, and we adopt the AdamW optimizer with an initial learning rate of 1e-5. The input image resolution is $1024 \times 1024$, with each clip containing 6 frames and supporting up to 10 object instances. We employ automatic mixed precision (AMP) and apply gradient clipping (max norm = 0.1) to stabilize convergence. The training is conducted on three NVIDIA H100 NVLink GPUs (each with 96GB memory) for 50 epochs, taking approximately 40 hours in total.

\subsection{Main Results}
\label{sec:main_results}

\begin{table}[ht]
\centering
\begin{tabular*}{\linewidth}{@{\extracolsep{\fill}}cccc}
\toprule
\textbf{Method} & \textbf{J\&F} & \textbf{J} & \textbf{F} \\
\midrule
imaplus (1st) & 0.8726 & 0.8359 & 0.9092 \\
\textcolor{black}{KirinCZW (ours)} & \textbf{0.8628} & \textbf{0.8250} & \textbf{0.9007} \\
dumplings (3rd) & 0.8392 & 0.8028 & 0.8757 \\
\bottomrule
\end{tabular*}
\caption{Leaderboard of the MOSE Track in the 4th PVUW Challenge 2025.}
\label{tab:leaderboard}
\end{table}

Our method achieved second place in the MOSE Track of the CVPR 2025 PVUW Challenge, with a J\&F score of \textbf{0.8628}, consisting of a region similarity (J) of \textbf{0.8250} and a contour accuracy (F) of \textbf{0.9007}. The leaderboard is summarized in \cref{tab:leaderboard}, demonstrating the robustness and effectiveness of our method in complex video object segmentation scenarios.

\subsection{Ablation Study}
\label{sec:ablation}

To evaluate the contribution of each component, we conduct ablation studies on the MOSE validation set. As shown in \cref{tab:ablation}, we start from the public baseline Cutie, which achieved a J\&F of 0.7065 on MOSE---significantly lower than its performance on traditional datasets like YouTube-VOS, indicating the challenges posed by complex scenes.

\begin{table}[ht]
\centering
\begin{tabular*}{\linewidth}{@{\extracolsep{\fill}}cccc}
\toprule
\textbf{Method} & \textbf{J} & \textbf{F} & \textbf{J\&F} \\
\midrule
Cutie (val) & 0.6511 & 0.7619 & 0.7065 \\
Baseline (val) & 0.6953 & 0.7761 & 0.7357 \\
Baseline + DA (val) & 0.7181 & 0.7947 & 0.7564 \\
Baseline + DA + MSS (val) & 0.7339 & 0.8191 & 0.7765 \\
Baseline + DA + MSS (test) & \textbf{0.8250} & \textbf{0.9007} & \textbf{0.8628} \\
\bottomrule
\end{tabular*}
\caption{Ablation study results on the MOSE validation and test sets.}
\label{tab:ablation}
\end{table}

We then evaluate our baseline framework with multi-scale encoders, enhanced memory, and mask mechanisms, which improves J\&F to 0.7357. Adding MOSE+ and our proposed data augmentation (DA) strategy further raises performance to 0.7564, verifying the effectiveness of enhanced training data. Finally, we introduce our Mask Scaling Strategy (MSS) to dynamically adjust the output mask distribution using \textit{sigmoid scale} and \textit{bias}, which boosts the performance to 0.7765 on the validation set and \textbf{0.8628} on the test set.As shown by the qualitative results in \cref{fig:output}, our optimized method demonstrates superior performance on the MOSE test set, particularly in segmenting small objects, occluded targets, and objects within complex and cluttered scenes.

\section{Conclusion}
\label{sec:conclusion}
In this work, we proposed a robust solution for complex video object segmentation by integrating enhanced training strategies and a mask confidence control mechanism. Based on the MOSE+ dataset, our approach incorporates targeted data augmentation and adaptive mask output calibration to improve segmentation performance under challenging scenarios. Our method achieved a J\&F score of 0.8628 and ranked second in the MOSE track of the CVPR 2025 PVUW Challenge.